\documentclass[10pt,twocolumn,letterpaper]{article}

\usepackage{iccv}
\usepackage{times}
\usepackage{epsfig}
\usepackage{graphicx}
\usepackage{amsmath}
\usepackage{amssymb}
\usepackage{balance}
\usepackage[pagebackref=true,breaklinks=true,letterpaper=true,colorlinks,bookmarks=false]{hyperref}
\usepackage{subfig}
\DeclareMathOperator*{\argmax}{argmax}

\iccvfinalcopy 

\def\iccvPaperID{} 

\ificcvfinal\fi

\begin{document}

\title{Show and Recall: Learning What Makes Videos Memorable}

\author{Sumit Shekhar\\
Adobe Research\\
{\tt\small sushekha@adobe.com}
\and
Dhruv Singal\\
Adobe Research\\
{\tt\small dhsingal@adobe.com}
\and
Harvineet Singh\\
Adobe Research\\
{\tt\small harvines@adobe.com}
\and
Akhil Shetty\\
University of California, Berkeley\\
{\tt\small shetty.akhil@berkeley.edu}
\and
Manav Kedia\\
University of Illinois at Urbana-Champaign\\
{\tt\small mkedia2@illinois.edu}
}

\maketitle

\begin{abstract}
With the explosion of video content on the Internet, there is a need for research on methods for video analysis which take human cognition into account. One such cognitive measure is memorability, or the ability to recall visual content after watching it. Prior research has looked into image memorability and shown that it is intrinsic to visual content, but the problem of modeling video memorability has not been addressed sufficiently. In this work, we develop a prediction model for video memorability, including complexities of video content in it. Detailed feature analysis reveals that the proposed method correlates well with existing findings on memorability. We also describe a novel experiment of predicting video sub-shot memorability and show that our approach improves over current memorability methods in this task. Experiments on standard datasets demonstrate that the proposed metric can achieve results on par or better than the state-of-the art methods for video summarization.
\end{abstract}

\section{Introduction} \label{sec:Intro}
Internet today is inundated with videos. The popular video site, YouTube, alone has more than a billion users and millions of hours of videos being watched every day \cite{YouTubeStat}. Thus, it has become imperative to investigate into advanced technologies for organization and curation of videos. Further, as any such system would involve interaction with humans, it is essential to take cognitive and psychological factors into account for designing an effective system. Moreover, it has been show that metrics like popularity~\cite{Khosla2014} and virality~\cite{Deza2015} can be predicted by analysing visual features.

An important aspect of human cognition is memorability or the ability to recall visual content after viewing it. Memorability is intricately related to perceptual storage capacity of human memory \cite{brady2008visual}. Recent studies have further shown that for prediction of image memorability, deep trained features can achieve near human consistency \cite{ICCV15Khosla}. There have been also related works in image memorability exploring different features and methods \cite{khosla2013modifying, kim2013relative, celikkale2013visual}. However, modeling and predicting memorability for video content has not been looked into sufficiently. This is a challenging problem because of added complexities of video like duration, frame rate, etc. Videos also convey multitude of visual concepts to the user, hence, it becomes difficult to ascertain the memorability of the overall content.  Further, the temporal structure of the video also needs to be taken into account while modeling video content memorability. 

 An earlier approach to model video memorability by Han \textit{et al.} \cite{han2015learning}  deploys a survey-based recall experiment. Here, the participants (about $20$) were initially made to watch several videos played together in a sequence, followed by a recall task after two days or a week, where they were asked if they remember the videos being shown. The score for a video was taken to be the fraction of correct responses by the participants. Due to the long time span of experiment, it is difficult to scale it to larger participant size. Further, there is no control over the user behavior between viewing and recall stage. Moreover, the method used fMRI measurements for predicting memorability, which would be difficult to generalize. 
 
 To this end, we design an efficient method to compute video memorability, which can be further generalized to applications like video summarization or search. The proposed framework required the participants to complete a survey-based recall task, where they initially watched several videos in a sequence, similar to the earlier approach. However, the recall experiment started after a short rest period of $30s$, and the participants were asked textual recall questions, instead of the full video being shown again. The textual questions were constructed from manual annotations of the videos. This was inspired from previous work in human memory research \cite{collins1969retrieval, baddeley1974working}, which showed that human memory stores information semantically. Further, the procedure of a textual questions-based recall survey has been followed in experimental psychology literature \cite{Jacoby1993}. The response time of the user was taken to be the measure of video memorability. Thus, the proposed survey avoids the long gap between viewing and recall stage, hence, making it scalable and efficient, as compared to \cite{han2015learning}. We will further release the video memorability dataset in public to help advance research in the field\footnote{\url{https://research.adobe.com/project/video-memorability/}}.
 
We conduct an extensive feature analysis to build a robust memorability predictor. We provide a baseline using state-of-the-art deep learning-based video features. We also explore semantic description, saliency and color descriptors, which have been found to be useful for memorability prediction in prior work on images \cite{dubey2015makes, ICCV15Khosla}. Further spatio-temporal features are added to describe video dynamics. We further show that the proposed video memorability model improves over static image memorability in predicting the memorability of video sub-shots ($3-5s$ video clips around a video frame). This experiment validates that image memorability is not sufficient to model memorability of video content. We demonstrate application of the model to video summarization task. Videos can be captured for different purposes, with diverse content, duration and quality. Video summarization is, therefore, a challenging task, especially for the content creators who want to ensure that the summary is remembered well by the viewers \cite{le2016impact}. In this work, we show that the proposed video memorability framework, which captures human memory recall, can further improve the state-of-the-art in video summarization.
The contribution of the work is as follows:\\	
1. We present a novel method for measuring video memorability through a crowd-sourced experiment.\\
2. We establish memorability as a valuable metric for video summarization and show better or at par performance with the state-of-the-art methods.\\ 
3. We demonstrate that proposed image memorability is not sufficient for analyzing memorability of short videos (or sub-shots).\\
4. We would further release an annotated video memorability dataset, to aid further research in the field.

\begin{figure*}[htp!]
\centering
\includegraphics[width=0.9\linewidth]{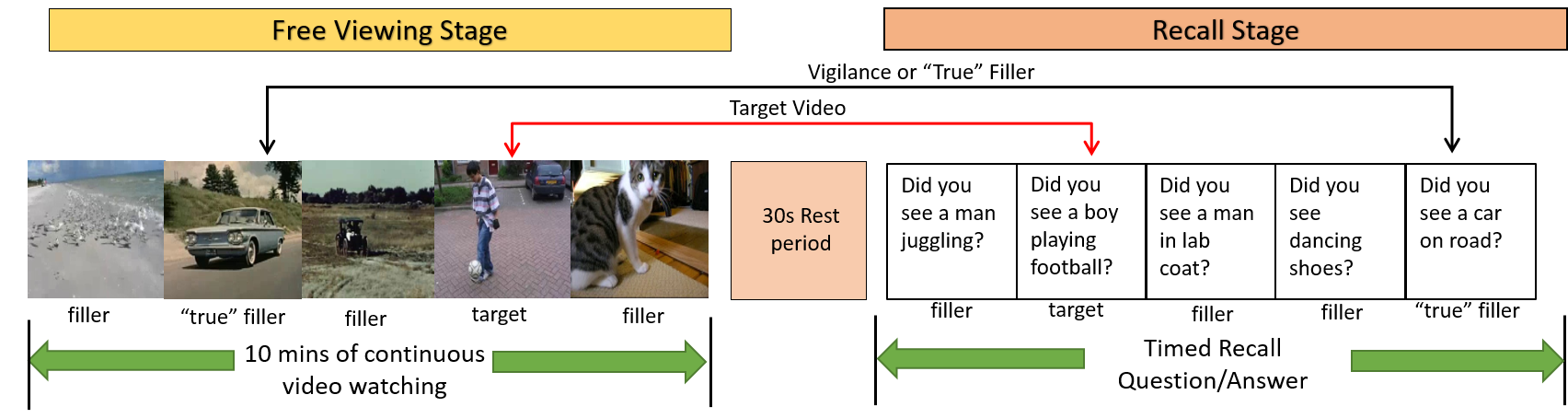}
\caption{Workflow for the proposed survey design to measure video memorability.}
\label{fig:MemSurveyWorkflow}
\end{figure*}

\section{Literature Survey} \label{sec:LitSurvey}
In this section, we discuss the prior work on memorability and related concepts of saliency and interestingness. We also describe the state-of-art work for video descriptors, video semantics and summarization.

\textbf{Memorability:} Recent works have explored memorability of images \cite{Isola2011, khosla2013modifying, kim2013relative, celikkale2013visual, ICCV15Khosla, isola2014makes}. Memorability of objects in images was studied in \cite{dubey2015makes}, while that of natural scenes was explored in \cite{mancas2013memorability}. There have been works studying the different aspects of memory, like visual capacity  \cite{standing1973learning, brady2008visual} as well as representation of visual scenes \cite{konkle2010scene}. The effect of extrinsic factors on memorability has also been looked into \cite{bylinskii2015intrinsic}. The recent work by Han \textit{et al.} \cite{han2015learning} models video memorability using fMRI measurements. 

\textbf{Saliency:} Saliency refers to the aspects of visual content which attract human attention.  There has been ample work in computing image saliency \cite{itti2005quantifying, toet2011computational, judd2009learning} and its applications to recognition tasks \cite{sharma2012discriminative}. The saliency feature has been found to be relevant for predicting memorability in \cite{dubey2015makes, ICCV15Khosla}.

\textbf{Interestingness:} Image interestingness was explored by Gygli \textit{et al.} \cite{gygli2013interestingness}, and was extended for video summarization task \cite{Gygli2015CVPR, gygli2014creating}. Interestingness score of an image was computed as a fraction of users who considered it interesting. However, the interestingness score is subjective and varies considerably with user preferences \cite{gygli2013interestingness}. Further, the applications to video summarization \cite{Gygli2015CVPR, gygli2014creating} use varied prediction models for interestingness. On the other hand, we demonstrate that the proposed memorability model can be generalized to different video summarization scenarios. Zen \textit{et al.} \cite{zen2016mouse} described an interestingness model using mouse activities. However, this may not generalize across different viewing conditions (e.g. mobile devices).

\textbf{Video descriptors:} Several deep learning-based features have been proposed for video classification task. There have been attempts at extending image-based deep features to videos using different fusion schemes \cite{karpathy2014large, yue2015beyond}. Tran \textit{et al.} \cite{tran2015learning} described a 3D-CNN model for action recognition in videos. Fusion of appearance and motion models using deep learning have also been explored \cite{simonyan2014two, feichtenhofer2016convolutional}. In addition, shallow features like  dense spatio-temporal \cite{wang2013action} have also been shown to improve classification accuracy when used in conjunction with deep learning features.

\textbf{Video semantics:} Inspired by the work on image captioning, there have been recent works in language description of videos. In particular, Donahue \textit{et al.} \cite{donahue2015long} proposed an LSTM-based approach for video description. This was followed by several works exploiting recurrent network architecture for describing videos \cite{venugopalan2015sequence, pan2016jointly, pan2016hierarchical}. Recently, a method exploiting external information for video captioning was described in \cite{venugopalan16emnlp}. 

\textbf{Video summarization:} There is rich literature on different video summarization techniques. Recently, there has been work exploring sub-modular optimization \cite{Gygli2015CVPR}, exemplars \cite{zhang2016summary}, object proposals \cite{Meng2016CVPR} and \textit{Determinantal point processes} (DPPs) \cite{zhang2016video} for summarization. There have also been related works in summarizing ego-centric videos \cite{lu2013story, lee2015predicting}. A keyword query-based summary method is described in \cite{sharghi2016query}.

\section{Video Memorability} \label{sec:VideoMem}
In this section, we describe modeling of video memorability, followed by a detailed analysis of memorability results. Figure~\ref{fig:MemSurveyWorkflow} shows the overall workflow for memorability ground truth collection.

\subsection{Ground Truth Collection}
The first step to model video memorability is to collect the ground truth of memorability scores. This was done through an Amazon Mechanical Turk (AMT) based crowd-sourced experiment on TRECVID 2012~\cite{trecvid2012} dataset.
 
\subsubsection{Dataset}
TRECVID 2012~\cite{trecvid2012} consists of about $500$ videos taken from the internet archives, ranging across various categories like nature, sports, animal and amateur home videos. The duration of the videos was typically $15-30s$. For the survey, the videos were first manually captioned to capture the content shown in them. Then, $116$ videos were selected across the various categories. These videos were partitioned into two sets - $100$ \emph{target} videos and $16$ \emph{filler} videos. The target videos were used to construct our model. For the survey, $25$ unique combinations of videos, each consisting of $4$ different target videos and the same set of $16$ filler videos were prepared. Thereafter, $4$ permutations of each of these $25$ combinations were created, keeping the order and the positions of the filler videos fixed. For the remaining positions, the order of target videos were changed according to the Latin square arrangement~\cite{winer1971statistical}. This ensured that each target video was shown at $4$ different positions to the users. The overall length of each of these $100$ video sequences was about $10$ minutes.

\subsubsection{Survey Design}
We conducted a recall-based experiment on AMT to collect the memorability ground truth. For the experiment, the participants had to first complete watching a full sequence, without browsing away from the survey page. To avoid any observer effect, the participants were not informed about the recall experiment at the end of free viewing. Further they were not allowed to repeat the survey.

After viewing the video sequence, there was a rest period of $30s$ and then, the subject was asked $20$ yes/no questions. He was given $5s$ to respond to each of them, and there was no provision of changing the response after the time was over. No response within the $5s$ duration was treated as a wrong reply. The questions were constructed from the manual text annotation for the video. Some sample questions from the survey are presented in Figure~\ref{fig:MemSurveyWorkflow}. Out of the $20$ questions, $8$ were true positives, out of which $4$ corresponded to the  target videos. The rest $4$ were randomly chosen from the $16$ filler videos, which we call as vigilance or "true" filler videos. The rest of the questions did not relate to any of the shown videos. The questions were randomly ordered for each survey to avoid any systematic bias in response. It was manually ensured that no two textual questions nor any two videos in a sequence were similar in content. The time that the subject took to respond each question was recorded. The survey was conducted with $500$ AMT workers, with each sequence permutation being viewed $5$ times, hence, giving $20$ responses for each target video.

\subsubsection{Memorability Score Computation}
The memorability score for each video was then calculated as follows:
\begin{itemize}
\item First, participants with precision less than $50 \%$ were removed from further calculations. This precision was calculated over both target and vigilance videos. This was done to remove users, who may have answered the questions in a random fashion (random behavior precision is about $40 \%$ for this setting). 
\item Consider a target video $i$ seen by participant $j$. Then, memorability score ($\mathit{MemScore}$) of the video $i$ for participant $j$ is:
\begin{align}
\mathit{\mathit{MemScore}(i,j)} & = \begin{cases} \frac{r(i,j)}{\bar{r}(j)} & \text{correct recall} \\
			 0 & \text{otherwise} \end{cases} 
\end{align}
where, 
$r(i,j)$ is the time left for participant $j$ in recalling video $i$, and
$\bar{r}(j)$ is the mean time left for participant $j$, in correctly recalling the videos (including filler videos). For incorrect responses time left was taken to be $0$. 
\item For the video $i$, the final memorability score, $\mathit{MemScore(i)}$, is the average score across all the participants:
\begin{equation}
\mathit{MemScore(i)} = \frac{\sum_{j=1}^N \mathit{MemScore(i,j)}}{N_i}
\end{equation}
where, $N_i$ is number of participants viewing the video.
\end{itemize}
Note that unlike the hit rate metric in \cite{ICCV15Khosla, isola2014makes}, we use a continuous metric based on recall time to capture the strength of memory, inspired from the work of Mickes \emph{et al.}~\cite{Mickes2007}. However, we do find that there is high correlation between hit rate and the proposed metric ($\rho$ = 0.91). Further we follow a user-based normalization for score calculation to neutralize background factors like the system used for answering the survey or biases specific to the user, which might affect the user response time.
  
\begin{figure*}[htp!]
\centering
\includegraphics[width=0.96\linewidth]{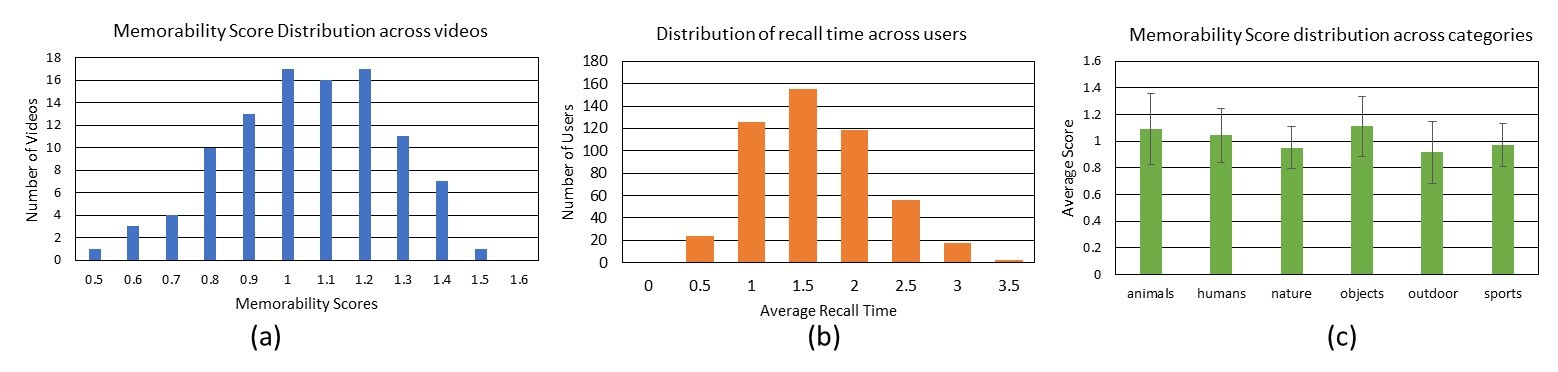}
\caption{Analysis of the output memorability scores: Distribution of (a) scores across videos, (b) recall time of participants and (c) scores over categories.}
\label{fig:MemScoreAnalysis}
\end{figure*}

\subsection{Memorability Analysis} \label{subsec:MemAnalysis}
Here, we analyze the output of the crowd-sourced survey for modeling video memorability. \\
\textbf{Memorability Score Distribution:} Figure \ref{fig:MemScoreAnalysis}(a) shows the distribution of video memorability scores across videos. It can be seen that the distribution peaks around $1$, while more memorable videos getting scores in the range of $1.3 - 1.5$. The overall distribution is skewed, with some videos getting scores as low as $0.5$. The high scores around $1.3 - 1.5$ means that more memorable videos are recalled faster than the average user recall time. Some of the least and the most memorable videos are shown in Figure \ref{fig:TopBottomMem}. \\  
\textbf{User Response Time:} Figure \ref{fig:MemScoreAnalysis}(b) shows that the  distribution of average user recall time has considerable variation. This justifies our choice of user-based normalization for score calculations. \\ 
\textbf{Effect of video category:} Average memorability scores for different categories of videos are shown in Figure \ref{fig:MemScoreAnalysis}(c). It can be seen that \textit{animals} and \textit{objects} videos are most memorable (also as per Figure \ref{fig:TopBottomMem}), followed by \textit{human} and \textit{sports} videos. The \textit{nature} and \textit{outdoor} videos have lower scores on an average. Thus, the semantic category of the video also affects its memorability.\\ 
\textbf{Human Response Consistency:} We also analyzed the consistency of human responses in the AMT survey. The output responses were divided randomly into equal halves, and Spearman's correlation ($\rho$) was calculated between the memorability score outputs of the two halves. The process was repeated $25$ times. We get a high average correlation, $\rho = 0.68$, which is consistent with findings in the previous works \cite{isola2014makes, dubey2015makes} that memorability is intrinsic to the visual content.\\
\textbf{Effect of complexity of textual questions:} The correlation of Flesch-Kincaid Grade Level readability metric~\cite{Kincaid1975} of the textual questions with the memorability scores was found to be quite low ($\rho = 0.0003$). Thus, we don't observe any effect of complexity of questions on memorability scores.

\section{Predicting Memorability} \label{sec:PredMem}
In this section, we discuss the task of predicting video memorability. The feature extraction from videos is described in Section \ref{subsec:featExtractPredict} and an analysis of features for memorability prediction is discussed in Section \ref{subsec:featAnalyzePredict}.

\subsection{Feature Extraction} \label{subsec:featExtractPredict}
Previous works on memorability \cite{dubey2015makes, ICCV15Khosla} have shown semantics, saliency and color to be important features for predicting memorability. Further, we extract spatio-temporal features to represent video dynamics, and provide a baseline using a recent, state-of-art deep learning feature for video classification.

\begin{itemize}
\item \textbf{Deep Learning (DL):} We extracted the recently proposed C3D deep learning feature \cite{tran2015learning}, trained on the Sports-1M dataset \cite{karpathy2014large} from the videos. The feature has been shown to achieve state-of-the-art classification results on different video datasets. Following the work, we used the activation of the $\mathit{fc-6}$ layer of the pre-trained C3D network to create a $4096$-dimensional representation of the video.

\item \textbf{Video Semantics (SEM):} We used the improved video captioning method developed in Venugopalan \textit{et al.} \cite{venugopalan16emnlp} to first generate the semantic description of the videos. The generated text was then fed to a recursive auto-encoder network \cite{socher2011semi} to generate a $100$-dimensional representation of the videos.  

\item  \textbf{Saliency (SAL):} Saliency or the aspect of visual content which grabs human attention has shown to be useful in predicting memorability \cite{isola2014makes, dubey2015makes}. We extracted the saliency feature for the video as follows. First, we generated saliency probability maps, using the method proposed in \cite{judd2009learning}, on $10$ frames extracted at uniform intervals from the video. This was followed by averaging the saliency maps over the frames, and re-sizing the averaged map to $50 \times 50$, followed by vectorization, to get the final feature.   

\item \textbf{Spatio-Temporal features (ST):} We used the recent state-of-the-art dense trajectory method \cite{wang2013action} to extract a $4000$-dimensional vector to represent the spatio-temporal aspect of the video.  

\item \textbf{Color features (COL):} A $100$-dimensional color feature was generated for each video by averaging the $50$-binned hue and saturation histograms for $10$ frames extracted at uniform intervals from the video, followed by concatenation. 
\end{itemize}

\subsection{Prediction Analysis} \label{subsec:featAnalyzePredict}
Here, we describe the training of regressor for predicting video memorability, and an analysis of importance of different features. For training the regressor, the dataset was randomly split into $80$ training videos and the rest $20$ for test, and the process was repeated $25$ times.  We used random forest (RF) regressor to train the model for individual features, tuned using cross-validation. For combining the features, we simply averaged the output regression scores of the individual features. Table \ref{tab:FeatCombPerform} reports RMSE for different feature combinations. The results are obtained by averaging over all the $25$ runs.

\begin{table}[htp!]
\centering
\begin{tabular}{|c|c|c|}
\hline
\textbf{Features} & \textbf{RMSE} \\ \hline
COL & $0.155 \pm 0.001$ \\ \hline
ST \cite{wang2013action} & $0.146 \pm 0.002$ \\ \hline
SAL \cite{judd2009learning} & $0.142 \pm 0.002$ \\ \hline
DL (C3D \cite{tran2015learning})  & $0.140 \pm 0.002$ \\ \hline
SEM \cite{venugopalan16emnlp} & $0.138 \pm 0.003$ \\ \hline
DL+ST+SAL+COL & $0.136 \pm 0.001$ \\ \hline
\textbf{SEM+ST+SAL+COL} & $\mathbf{0.135 \pm 0.001}$ \\
\hline
\end{tabular}
\caption{Performance analysis of different features. DL: Deep Learning, SEM: Semantics, SAL: Saliency, ST: Spatio-temporal, COL: Color.}
\label{tab:FeatCombPerform}
\end{table}

\begin{figure*}[htp!]
\centering
\includegraphics[width=0.82\linewidth]{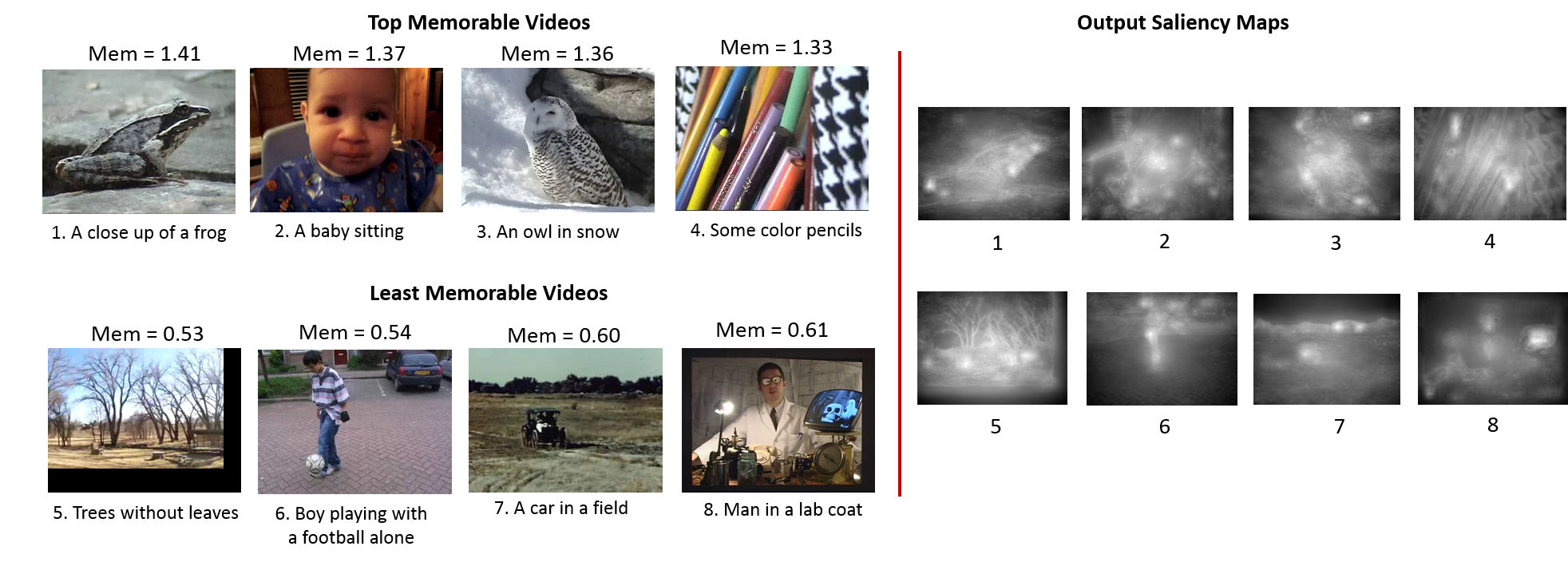}
\caption{Examples of the most and the least memorable videos from the crowd-sourced video memorability experiment with TRECVID 2012. The output saliency maps \cite{judd2009learning} for these videos are displayed along with.}
\label{fig:TopBottomMem}
\end{figure*}

\textbf{Feature Analysis:} Table \ref{tab:FeatCombPerform} shows the performances for different feature combinations. It can be seen that the deep learning-based features, DL and SEM individually achieve low RMSE values, with the latter performing better. Among the shallow features, SAL feature exhibit the lowest RMSE followed by ST and COL features. The better performance of SAL features might be because it captures if the subject of the video grabs human attention or not, as seen in Figure \ref{fig:TopBottomMem}. It can be seen that the top $3$ memorable videos have salient foreground. However, the \textit{color pencils} video, having similar saliency map as the \textit{tree without leaves} video, has a very different memorability score. Thus, saliency alone is not sufficient to explain memorability. The worse performance for ST might be because video dynamics alone is not sufficient to account for the memorability score. Overall feature combinations further lower the RMSE values. \\

\textbf{Final Memorability Predictor:} The final memorability classifier was trained over all of the $100$ target videos, using the SEM+ST+SAL+COL feature combination. We used this regressor for all further experiments. 

\section{Sub-Shot Memorability} \label{sec:SubShot}
In this section, we discuss the problem of predicting sub-shot memorability, and how the existing image memorability work is not sufficient to address the same. We define a sub-shot as a short clip of around $3-5$s around a selected frame of the video. Due to the short duration, the sub-shot can generally be considered to have homogeneous composition, and that the selected frame is a good representation of the sub-shot. We conducted a survey-based recall experiment to collect the memorability ground truth for sub-shots, following the procedure as described in Section \ref{sec:VideoMem}.

First, we selected $50$ target videos from the TRECVID 2012 \cite{trecvid2012}, different from the target videos used in Section \ref{sec:VideoMem}. For each video, a sub-shot of $3$s around an image frame selected randomly, was extracted. The crowd-sourced AMT survey was designed following the procedure described for video memorability in Section \ref{sec:VideoMem}, except for a change at the recall stage. 

The free viewing sequences consisted of $4$ target sub-shots and $16$ filler sub-shots, as before. The filler sub-shots were extracted from the filler videos used for the earlier experiment. Due to the shorter video lengths, total free viewing stage lasted for around $2$ minutes. During the recall experiment, the participants were asked if they can recall the displayed images (instead of the textual question in original survey). This was done because the sub-shot can be represented by the chosen image frame effectively. For the target as well as ``true" filler sub-shots, the corresponding image frame of the video was used, for other slots random images corresponding to none of the shown videos were used. The images were flashed for $0.5$s, and then the subject was asked if he can recall the displayed image in $5s$. The final score was calculated using the procedure described in Section \ref{sec:VideoMem}.

\textbf{Human Response Consistency:} A consistency analysis of the annotations, similar to the one conducted in Section \ref{subsec:MemAnalysis} yielded a Spearman's correlation of $0.45$. Thus, sub-shots also have consistent memorability across users, similar to video memorability.

\textbf{Prediction Analysis:} We conducted a comparison of the proposed video memorability regressor with the existing image memorability work \cite{ICCV15Khosla}, on predicting the memorability of sub-shots. Image memorability scores were computed by running the pre-trained model from \cite{ICCV15Khosla} on the selected frame for each sub-shot. Table \ref{tab:SubshotMemScore} 
demonstrates the results of the comparison. It can be seen that image memorability yields much lower Spearman's correlation value than video memorability regressor. This result demonstrates that complexities of video data must be accounted for, in order to predict memorability. Further, the moderate-to-low correlation for both the cases indicate that further investigations are required into how memorability predictions can be generalized across different kinds of tasks (\textit{e.g.} video to sub-shot or image to sub-shot). 

\begin{table}[htp!]
\centering
\begin{tabular}{|c|c|c|}
\hline
\textbf{Method} & \textbf{Spearman's cor. ($\rho$)} \\ \hline
Image Mem. \cite{ICCV15Khosla} & $0.06$ \\ 
Video Mem. & $0.20$ \\
\hline
\end{tabular}
\caption{Correlation results for image memorability \cite{ICCV15Khosla} and the proposed video memorability with the ground truth.}
\label{tab:SubshotMemScore}
\end{table}

\section{Video Summarization} \label{sec:VideoSummary}
In this section, we describe the application of the proposed method to video summarization tasks. Recently a state-of-the-art algorithm for summarization based on supervised learning of sub-modular objective function was proposed by Gygli \textit{et al.} \cite{Gygli2015CVPR}. The framework combined several image-based objectives like interestingness, uniformity and representativeness to improve the quality of video summary. The weights given to each of the objective criteria were learned using a supervised learning algorithm trained using reference human summaries. Here, we further incorporate the proposed video memorability framework as an objective criterion for summarization. We believe this would help improve quality of summaries further. \\

\textbf{Memorability objective:} For a video $\mathcal{V}$ partitioned into $N$ segments, $\{\mathbf{s}_i \}^N_{i=1}$, memorability objective, $\mathit{VidMem}$ for a selection of $K \subset \{ 1, 2, \cdots, N \}$ segments is defined as:
\begin{equation}
\mathit{VidMem} = \sum_{i \in K} \mathit{MemScore}(\mathbf{s}_i), 
\end{equation}
where, $MemScore(\mathbf{s}_i)$ is the predicted memorability score for segment $\mathbf{s}_i$. It can be shown that the objective function is sub-modular. Given the functions for scoring summaries on memorability ($\mathit{VidMem}$), uniformity ($\mathit{VidUnif}$)~\cite{Gygli2015CVPR} and representativeness ($\mathit{VidRep}$)~\cite{Gygli2015CVPR}, the overall objective criteria for selecting the summary, $\textbf{y}$ is given as:
\begin{equation} \label{eq:vsumm}
\textbf{y}_{opt} = \argmax\limits_{\textbf{y} \in 2^\mathcal{V}, |\textbf{y}| \leq L} \sum_{f \in \mathcal{F}} w_{f} f(\textbf{y}; \mathcal{V})
\end{equation}
where, $L$ is the length of summary, $\mathcal{F}:=\{\mathit{VidMem,VidRep,VidUnif}\}$, $f(\textbf{y}; \mathcal{V})$ is the summary score using $f$ and weights $w$ are learned using the supervised sub-modular optimization as described in \cite{Gygli2015CVPR}. The results are demonstrated on SumMe user \cite{gygli2014creating} and UT Egocentric \cite{Lee2012CVPR} video datasets.

\subsection{User Video Dataset}
The SumMe user video dataset \cite{gygli2014creating} consists of $25$ short videos with lengths ranging from $1-7$ minutes. The videos depict various activities like sports, cooking, different outdoor activities, traveling, etc. Each video has around $15$ (total $390$)  reference ground truth summaries, generated by humans in a controlled environment. We followed the pre-processing and evaluation protocol described in \cite{Gygli2015CVPR} to have a consistent comparison with the prior art.

\textbf{Pre-processing:} The videos were partitioned using super-frame segmentation method \cite{gygli2014creating}. For each segment, SEM, ST, SAL and COL features were extracted, and memorability scores were predicted using the final model (SEM+ST+SAL+COL) trained in Section \ref{sec:PredMem}. 

\textbf{Evaluation:} The dataset was split $12$-ways and a leave-one-out method was followed for evaluating the algorithm. The results were averaged over $100$ runs. The methods were evaluated for a budget of $15 \%$ of the extracted segments. The training was done using the reference user summaries for each video. During the test time, the generated summary was compared with all the reference summaries, and the maximum overlap was taken to get the final F-measure and Recall results, as described in \cite{gygli2014creating}. 

\begin{figure*}[htp!]
\centering
\includegraphics[width=0.92\linewidth]{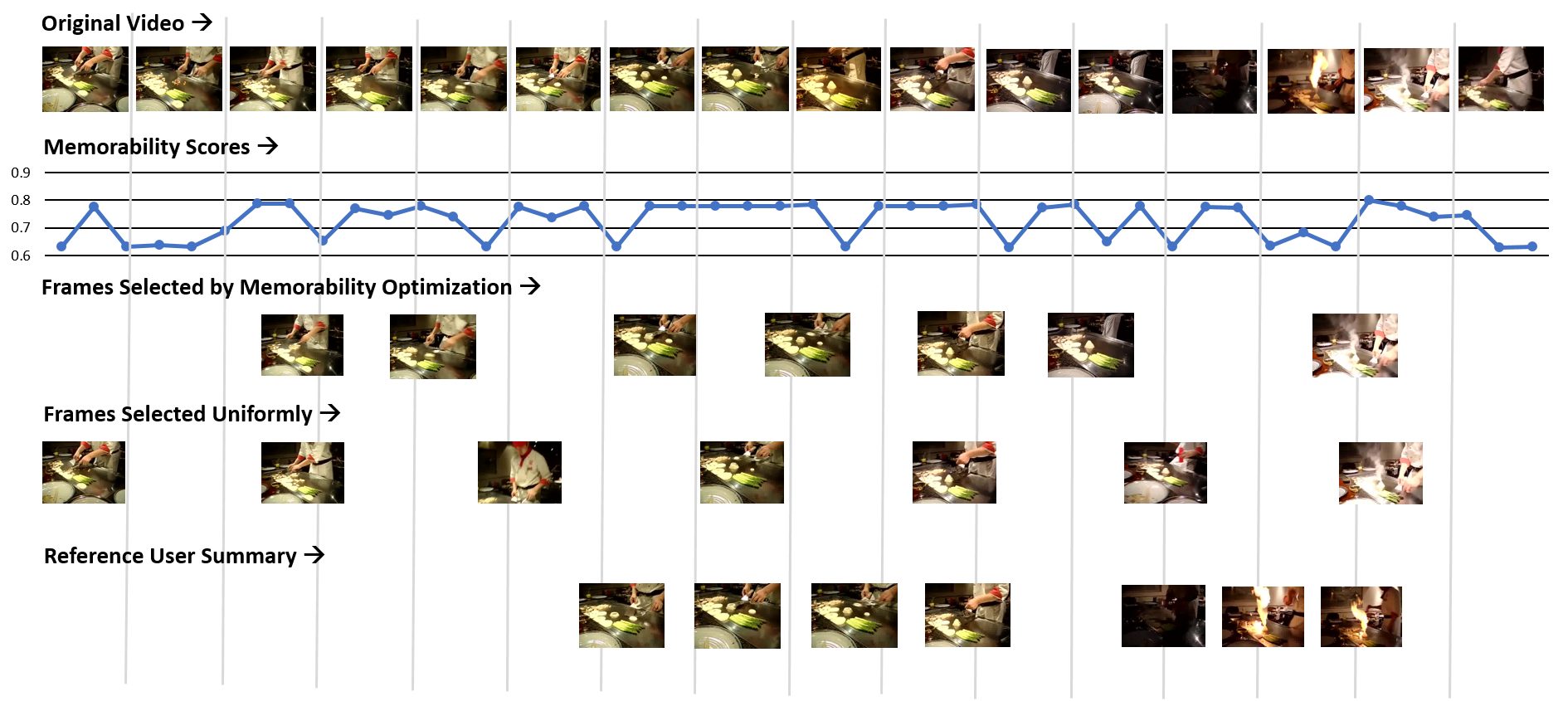}
\caption{An example of the frame selection through memorability criterion. The shown video from SumMe dataset has a cooking activity going on. As seen in the figure, compared to a uniform selection of frames, memorability criterion picks up frames more relevant to the reference user. It can been seen that memorability score can capture different events and transitions in the video.}
\label{fig:VideoSummaryExample}
\end{figure*}

\textbf{Results:} Table \ref{tab:EvalSumMe} shows the comparison of the proposed memorability-based framework with the previous methods. It can be seen that the proposed video memorability alone is able to achieve state-of-the-art F-measure score on the dataset. The results further increase through combination with representativeness and uniformity objectives. Further, the memorability objective gets $96 \%$ weight in the supervised training with all the objectives, thus, reinforcing the usefulness of the method. Further, an illustration of summarization achieved by using memorability objective is shown in Figure \ref{fig:VideoSummaryExample}. It can be seen that memorability picks up frames more relevant to users, as well as captures different events in videos well.

\begin{table}[htp!]
\centering
\begin{tabular}{|c|c|c|}
\hline 
\textbf{Methods} & \textbf{F-measure} & \textbf{Recall} \\ \hline
UserSum \cite{gygli2014creating} & $39.34 \pm 0.00 \% $ & $\mathbf{44.44 \pm 0.00 \%}$ \\
Uniformity & $24.68 \pm 0.04 \%$ & $27.08 \pm 0.08 \%$\\ 
Representativeness & $26.69 \pm 0.00 \%$ & $26.65 \pm 0.00 \%$ \\ 
Interesting \cite{Gygli2015CVPR} & $39.52 \pm 0.00 \%$ & $42.50 \pm 0.00 \%$ \\
Uni.+Rep.+Int. \cite{Gygli2015CVPR} & $39.68 \pm 0.09 \%$ & $43.01 \pm 0.08 \%$ \\
Zhang \textit{et al.} \cite{zhang2016summary} & $40.9 \%$ & $-$ \\
Vid. Memorability & $41.11 \pm 0.10 \%$ & $37.91 \pm 0.11 \%$ \\
Uni.+Rep.+Mem. & $\mathbf{41.21 \pm 0.12 \%}$ & $38.41 \pm 0.15 \%$ \\ \hline 
\end{tabular}
\caption{Evaluation results for summarization with $15 \%$ budget on SumMe dataset.}
\label{tab:EvalSumMe}
\end{table}

\subsection{UT Egocentric (UTE) Dataset} 
UTE dataset \cite{Lee2012CVPR} consists of $4$ videos, each with $3-5$ hours of video content. The video content was recorded through a wearable camera, and logs day activities of the wearer. Thus, the videos may be repetitive and were shot in an uncontrolled fashion. Textual captions for $5s$ segments of each of the video, as well as $3$ reference summaries for each video were provided by Yeung \textit{et al.} \cite{yeung2014videoset}. We followed the pre-processing and evaluation protocol described in \cite{Gygli2015CVPR} to have a consistent comparison with the prior art. 

\textbf{Pre-processing:} The videos were divided into $5$s segments and then memorability score was calculated for each segment as described in the previous experiment. For each segment, SEM, ST, SAL and COL features were extracted, and memorability scores were predicted using the final model (SEM+ST+SAL+COL) trained in Section \ref{sec:PredMem}. 

\textbf{Evaluation:} Firstly, for all the videos, segment-based reference summaries were generated using the provided textual summary, following the method proposed in \cite{yeung2014videoset}. We used greedy optimization based on bag-of-word model to produce the segment-based reference summaries. The dataset was then split $4$-ways and a leave-one-out train/test was followed, similar to \cite{Gygli2015CVPR}. The results were averaged over $100$ runs. During the test time, the generated segment-based summary was converted to textual description, and then was compared to the reference text summaries using ROUGE-SU method \cite{lin2004rouge}.  The ROUGE-SU computes unigram and skip-bigram co-occurence between candidate and reference summaries, after stemming and removing stop word in the summaries. 

\textbf{Results:} The proposed method was evaluated for two summary lengths - a shorter length of $1$ min $20s$ and a longer length of $2$ mins. The results for evaluations are shown in Table \ref{tab:EvalEgoCentricShort} and Table \ref{tab:EvalEgoCentricLong}. It can be seen that the proposed method achieves the best recall for shorter summary length, whereas for longer summaries, the performance is comparable to Interestingness metric \cite{Gygli2015CVPR}. This may be because we do not employ the manual annotations provided in \cite{Lee2012CVPR} to identify important objects, which was used in interestingness calculation \cite{Gygli2015CVPR}. Further, with the increase in summary lengths, other metrics like uniformity and representativeness also give results close to memorability. This might be because in typical ego-centric videos, there would be only few ``memorable" segments relevant to user. With the increased budget other metrics can also capture these segments.  We believe that the memorability results could be further improved through enhancements in feature design. 

\begin{table}[htp!]
\centering
\begin{tabular}{|c|c|c|}
\hline 
\textbf{Method} & \textbf{F-measure} & \textbf{Recall} \\ \hline
Lee \textit{et al.} \cite{Lee2012CVPR} & $17.40 \pm 4.07 \%$ & $12.20 \pm 3.30 \%$ \\
Video MMR \cite{li2010multi} & $17.73 \pm 0.00 \%$ & $12.49 \pm 0.00 \%$ \\
Uniformity & $18.75 \pm 1.36 \%$ & $12.92 \pm 1.11 \%$\\ 
Representativeness & $19.08 \pm 0.00 \%$ & $12.95 \pm 0.00 \%$ \\ 
Interesting \cite{Gygli2015CVPR} & $20.93 \pm 0.00 \%$ & $15.15 \pm 0.00 \%$ \\
Uni.+Rep.+Int. \cite{Gygli2015CVPR} & $\mathbf{21.91 \pm 0.06 \%}$ & $15.73 \pm 0.04 \%$\\
Vid. Memorability & $18.13 \pm 0.08 \%$ & $15.55 \pm 0.04 \%$ \\
Uni.+Rep.+Mem. & $19.37 \pm 0.08 \%$& $\mathbf{17.9 \pm 0.09 \%}$\\ \hline  
\end{tabular}
\caption{Evaluation results for shorter summarization ($1$ min $20s$) on UTE dataset.}
\label{tab:EvalEgoCentricShort}
\end{table}

\begin{table}[htp!]
\centering
\begin{tabular}{|c|c|c|}
\hline 
\textbf{Method} & \textbf{F-measure} & \textbf{Recall} \\ \hline
VideoMMR \cite{li2010multi} & $25.57 \pm 0.00 \%$ & $23.10 \pm 0.00 \%$ \\
Uniformity & $25.41 \pm 1.35 \%$ & $22.27 \pm 1.56 \%$ \\ 
Representativeness & $27.02 \pm 0.00 \%$ & $23.51 \pm 0.00 \%$ \\ 
Interesting \cite{Gygli2015CVPR} &$27.07 \pm 0.00 \%$ & $24.78 \pm 0.00 \%$\\
Uni.+Rep.+Int. \cite{Gygli2015CVPR} & $\mathbf{29.01 \pm 1.18 \%}$ & $\mathbf{26.21 \pm 1.23 \%}$ \\
Vid. Memorability & $26.81 \pm 0.04 \%$ & $21.24 \pm 0.02 \%$\\
Uni.+Rep.+Mem. & $28.3 \pm 0.5 \%$ & $23.6 \pm 0.3 \%$ \\ \hline 
\end{tabular}
\caption{Evaluation results for longer summarization ($2$ min) on UTE dataset.}
\label{tab:EvalEgoCentricLong}
\end{table}

\section{Conclusions and Future Work} \label{sec:Conclude}
In this work, we have described a robust way to model and compute video memorability. The computed memorability scores are consistent, hence, are intrinsic to the video content, as has been established by prior work in memorability. Further, we analyze different features in predicting memorability, and demonstrate importance of different features. A novel experiment on sub-shot memorability proves that image memorability alone is not sufficient to explain the memorability of sub-shots. Finally, the proposed method achieves state-of-the-art results on different video summarization datasets. This shows that memorability is a viable criteria for creating extractive video summaries. In future, we plan to conduct the video memorability experiment on a larger scale, as well as, design improved features for prediction. This would also require methods proposed in crowd-sourcing literature for addressing ambiguity in the questions and labels \cite{Bragg2013,Deng2014}. We further believe that application of video memorability to challenging tasks, like video-based recognition or segmentation would enhance the current state-of-the-art.

\section{Acknowledgement}
We thank the anonymous reviewers for their feedback. We also thank our colleagues from Adobe Research who provided insight and expertise that greatly assisted the research. We particularly thank Atanu Ranjan Sinha and P. Anandan for their valuable inputs. 
{\small
\bibliographystyle{ieee}
\balance
\bibliography{paper_bib}
}

\end{document}